\documentclass{article}

\usepackage{arxiv}
\usepackage[utf8]{inputenc} 
\usepackage[T1]{fontenc}    
\usepackage{hyperref}       
\usepackage{url}            
\usepackage{booktabs}       
\usepackage{tabularx}       
\usepackage{seqsplit}       
\usepackage{amsfonts}       
\usepackage{nicefrac}       
\usepackage{microtype}      
\usepackage{lipsum}		
\usepackage{graphicx}
\usepackage{authblk}
\usepackage{doi}

\usepackage[
backend=biber,
style=chem-acs,
]{biblatex}
\title{A Modular Agentic Framework for Synthetically Constrained
Multi-Objective Hit-to-Lead Optimization}

\date{}
\author[1,2]{Kelvin P. Idanwekhai}
\author[2]{Enes Kelestemur}
\author[2]{Benjamin Strickland}
\author[2,3]{Matthew Hart}
\author[4]{Steini Davidsson}
\author[4]{Angelos Angelopoulos}
\author[4]{Ron Alterovitz}
\author[2]{Marcello DeLuca}
\author[2]{Alexander Tropsha}

\affil[1]{Department of Chemistry, UNC Chapel Hill, Chapel Hill, NC, USA}
\affil[2]{Division of Chemical Biology and Medicinal Chemistry, Eshelman School of Pharmacy at UNC, Chapel Hill, NC, USA}
\affil[3]{Department of Applied Physical Sciences, UNC Chapel Hill, NC, USA}
\affil[4]{Department of Computer Science, UNC Chapel Hill, Chapel Hill, NC, USA}

\affil[*]{Corresponding authors: \texttt{\{delmar,alex\_tropsha\}@unc.edu}}

\usepackage{amsmath}
\begin{document}
\maketitle


\begin{abstract}
Hit-to-lead optimization requires iterative design of hit analogs across
competing potency, selectivity, physicochemical, pharmacokinetic,
safety, and synthetic constraints. We present SABLE
(Synthetically-accessible Agentic Bayesian Ligand Exploration), an
open-source framework that employs natural-language orchestration to
guide chemical structure optimization. SABLE uses an LLM to interpret
user-defined goals and route tasks, while specialized tools perform
reaction-templated analog enumeration, physicochemical and ADMET
property prediction, structure-based affinity scoring, and Bayesian
optimization. The resulting workflow is a computational twin of the
analytical and prioritization stages of the design--make--test--analyze
cycle, providing provenance of each numerical output. Across single, and
multi-objective optimization studies, SABLE enriches candidate sets for
user-defined computational objectives while evaluating only a subset of
the enumerated search space. Its modular architecture allows tools and
characterization backends to be replaced by editing a simple config
file, without modifying operational logic. SABLE provides an extensible
decision-support framework for prioritizing synthetically constrained
analogs in early-stage drug discovery.
\end{abstract}

\section{Introduction}
Hit-to-lead optimization in medicinal chemistry proceeds through
iterative design--make--test--analyze (DMTA) cycles in which medicinal
chemists design, synthesize, and evaluate analogs of initial hits. The
objective is not simply to maximize target potency, but to develop a
chemical series that combines potency and selectivity with suitable
physicochemical properties, pharmacokinetics, safety, and synthetic
tractability. Each experimental optimization round expands the
structure--property relationships that inform the next set of
compounds\cite{schneiderRethinkingDrugDesign2019}. However, finite
synthetic and assay capacity limits evaluation to only a small fraction
of the accessible chemical analog space, often requiring multiple
chemical series to be advanced in parallel. Candidate selection must
therefore account for both the immediate benefit of a structural
modification and its compatibility with continued optimization. The
latter challenge is further complicated by the need to optimize
molecular structure not only for efficacy but also safety and drug-like
properties that contribute substantially to clinical
attrition\cite{sunWhy90Clinical2022a}.

This chemical structure optimization task is inherently multi-objective,
and it is further complicated by strong interdependency between
molecular properties. A structural modification intended to improve
potency may also alter selectivity, lipophilicity, solubility,
permeability, metabolic stability, or toxicity, and these effects cannot
always be inferred independently. For example, matched molecular-pair
analyses have shown that the same local transformation can produce
different, and sometimes opposing, effects on hERG inhibition,
solubility, and lipophilicity depending on its surrounding molecular
context\cite{papadatosLeadOptimizationUsing2010}. At the same time,
potency, absorption, distribution, metabolism, excretion, and safety
requirements frequently conflict at the campaign
level\cite{segallMultiParameterOptimizationIdentifying}. Historical
analyses have additionally cautioned that potency gains are often
accompanied by increases in molecular size or lipophilicity, a tendency
described as molecular inflation or molecular
obesity\cite{hannMolecularObesityPotency2011}. Thus, as stated above,
the central challenge is to develop multi-objective optimization
strategies that can systematically balance competing requirements and
guide compound selection toward an improved overall profile without
introducing liabilities that must be corrected in subsequent rounds.

Machine learning methods are well suited to address this computational
chemistry optimization challenge because they can evaluate large
candidate sets consistently across multiple objectives. For instance,
generative machine learning employing latent-space molecular generators,
sequence-based SMILES generation, and LLM-driven molecular
design\cite{grisoniBidirectionalMoleculeGeneration2020,pangDeepGenerativeModels2024,popovaDeepReinforcementLearning2018}
methods can propose large numbers of chemically valid structures with
the desired properties\cite{gaoSynthesizabilityMoleculesProposed2020}.
Such methods can be devised not merely to generate plausible molecules,
but to define an experimentally relevant search space and prioritize
candidates within it according to explicit, competing objectives.

Many of the components required to construct such a workflow are readily
available. Reaction-aware generation and enumeration methods can
restrict molecular design to structures assembled from specified
building blocks through encoded
transformations\cite{gaoGenerativeAINavigating2025,kelestemurHEALERHitExpansion}.
Property-prediction platforms and cheminformatics models provide
calculated descriptors or model-based estimates for physicochemical,
pharmacokinetic, and toxicological
endpoints\cite{dainaSwissADMEFreeWeb2017a,fuADMETlab30Updated2024,wellnitzSTOPLIGHTHitScoring2024b}.
Structure-based models such as Boltz-2 jointly predict biomolecular
complexes and affinity-related scores from protein-sequence and ligand
inputs\cite{passaroBoltz2AccurateEfficient2025a}. Bayesian optimization
(BO), in turn, provides a principled framework for sequential
decision-making when characterization is expensive and the relationship
between structure and optimization objective is
unknown\cite{frazierTutorialBayesianOptimization2018}. By balancing
exploration of uncertain regions with exploitation of candidates
predicted to perform well, BO provides a natural mechanism for
navigating a chemically constrained, multi-objective search space. The
remaining difficulty is therefore less the absence of computational
capabilities but the lack of a unified mechanism for coordinating them.
At present, these capabilities remain distributed across standalone web
services\cite{dainaSwissADMEFreeWeb2017a,fuADMETlab30Updated2024}, and
locally installed cheminformatics
toolkits\cite{jungVSFlowOpensourceLigandbased2023,leeChemalotChemalot_knimeCommand2017,oboyleOpenBabelOpen2011}.
Although individually capable, these resources demand nontrivial
expertise to install individual modules, integrate them into a pipeline,
and interpret the results of calculations, making such assembly, or a
toolkit, difficult to use by the practicing medicinal chemist without
dedicated informatics support. Any integrated system, therefore, must
fulfill two distinct demands: it must be usable by the medicinal
chemists whose design needs the tools are meant to support; and it must
remain open to advances that will supersede today\textquotesingle s
components. AutoLead proposed a framework that combines LLMs and BO for
lead optimization, but this workflow is limited by the modularity and
type of oracle used\cite{zhangAutoLeadLLMGuidedBayesian2026}. The
coverage of endpoints that can optimized by AutoLead, only include
descriptors computed by RDKit\cite{RDKit}, not accounting for other
ADMET and potency properties. In addition, the framework lacks two
important considerations: synthesizability-aware recommendation of
molecules and modularity of workflow. The emergence of standardized
model--tool interfaces, including the Model Context
Protocol\cite{WhatModelContext}, provides a potential technical
foundation for connecting modular software components and enabling the
incorporation of new tools as they become available. Standardized
connectivity alone, however, does not determine how chemical hypotheses
should be evaluated, how conflicting objectives should be reconciled, or
which candidate should be selected for the next iteration, reinforcing
the need for systems that are both accessible to practitioners and
adaptable to evolving computational capabilities.

Agentic methods offer one approach to this coordination problem by
allowing a language model to interpret a task and invoke appropriate
external tools. ChemCrow was, arguably, the first system that
established the value of grounding LLM responses in expert-designed
cheminformatics tools, with chemist evaluators preferring tool-augmented
outputs on nontrivial tasks\cite{m.branAugmentingLargeLanguage2024a}.
Coscientist extended this paradigm to robotic
synthesis\cite{boikoAutonomousChemicalResearch2023}, while ChemReasoner
combined Monte Carlo Tree Search (MCTS) with LLM-guided exploration for
catalyst design\cite{sprueillChemReasonerHeuristicSearch2024}. These
recent examples demonstrated that language models can provide accessible
interfaces for coordinating specialized chemical tools. They also raise
a separate question regarding the appropriate division of function
between the language model and the quantitative models it invokes. In
these architectures, the LLM remains central to scientific planning or
hypothesis generation, while external tools supply information,
calculations, or experimental execution. This arrangement can become
problematic when the language model is also expected to judge the
correctness of those outputs or choose among quantitatively competing
candidates. ChemCrow\textquotesingle s own evaluation illustrates the
cost. According to the authors of ChemCrow, when their evaluator LLM,
powered by GPT-4, was asked to assess completions, it could not reliably
distinguish clearly incorrect LLM outputs from tool-grounded
responses\cite{m.branAugmentingLargeLanguage2024a}. This finding
illustrates the potential risk of using language-model judgment as the
sole basis for quantitative assessment of chemical output, which has
also been echoed by previous works that evaluate the effectiveness of
LLMs as oracles in chemical optimization
tasks\cite{akkeBayesianOptimizationBiochemical,kristiadiSoberLookLLMs2024}.

\begin{figure}
	\centering
	\includegraphics[width=1\linewidth]{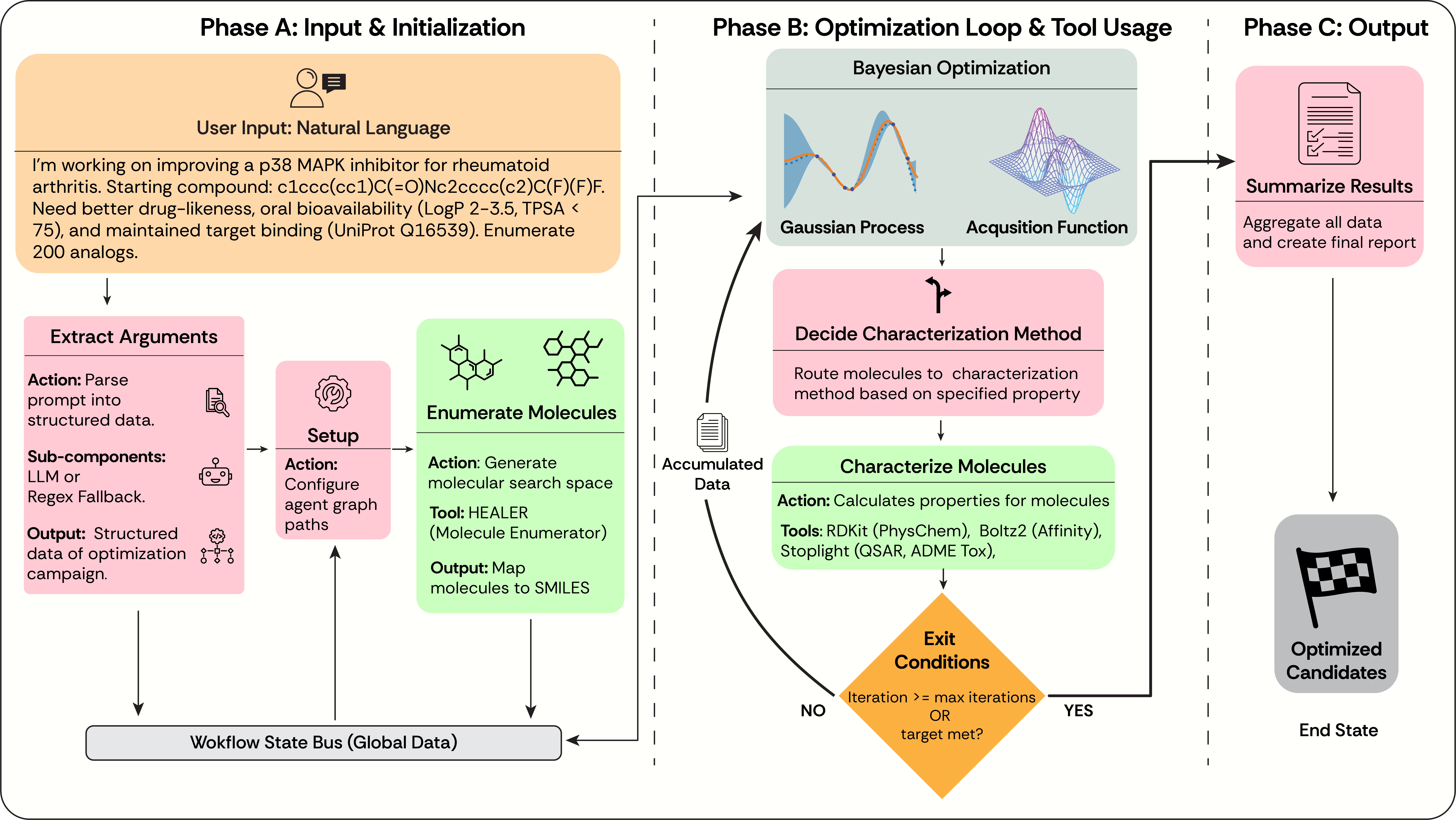}
	\caption{
Schematic representation of the ligand
optimization workflow. (A) A user prompts the system in natural
language; then, key arguments for the optimization run, like the
starting molecule, protein target, optimization objectives, and number
of iterations, are extracted with LLMs. Analogs of the starting compound
are also enumerated according to the user specification or default
configuration. (B) An optimization campaign is initiated on
enumerated compounds. Properties predicted with tools such as Boltz,
RDKit, or STOPLIGHT are used in place of experimental measurements. This
iteration is performed until predefined process termination criteria are
met (see Section 4.7). (C) Optimized candidates are selected
and reported to the user.}
	\label{fig:fig1}
\end{figure}

Herein, we introduce SABLE (Synthetically-accessible Agentic Bayesian
Ligand Exploration), a modular platform for early-stage hit-to-lead
optimization through language-model orchestration of specialized modules
supporting chemical design and optimization tasks (Fig. \ref{fig:fig1}). The LLM
interprets user-defined goals, extracts the parameters required to
configure the campaign, invokes the appropriate tools, and organizes
their outputs. Molecular enumeration, descriptor calculation, ADMET
property prediction, optimization, affinity scoring, and candidate
ranking are delegated to explicitly defined cheminformatics components.
Starting from a user-provided seed compound, SABLE uses Hit Expansion to
Advanced Leads using Enumerated Reactions (HEALER) to construct a
reaction-constrained analog space from available building blocks\cite{kelestemurHEALERHitExpansion}
(Fig. \ref{figs:figs1}), evaluates selected candidates using physicochemical, ADMET,
and structure-based characterization tools including
Boltz-2\cite{passaroBoltz2AccurateEfficient2025a}, and applies BO to
identify candidates that improve or balance the specified objectives.
The resulting workflow implements a computational analogue of the
analytical and prioritization stages of the DMTA cycle but is intended
to support, rather than replace, medicinal-chemistry review and
experimental validation.

By separating natural-language orchestration from candidate molecule
generation, quantitative assessment, and selection, SABLE makes the
optimization objectives, originating tools, and selection criteria
explicit and traceable. Its modular architecture further allows
enumeration, optimization, characterization, and experimental backends
to be incorporated or replaced without altering the central
orchestration and optimization logic. SABLE leans towards a bring your
own tools (BYOT) approach, allowing users to define their own tools and
predictors in a simple configuration file. More broadly, SABLE
establishes a general design pattern for scientific agents in which
language models provide an accessible and user-friendly interface
between researchers and computational tools, while domain-specific
models and explicit optimization algorithms retain responsibility for
quantitative scientific decisions.

\section{Results and Discussion}

\subsection{Workflow execution and state management}

SABLE operationalizes the architecture shown in Fig. \ref{fig:fig1} as a stateful
workflow implemented using LangGraph\cite{LangGraphOverview} (Fig. \ref{figs:figs2}).
Each campaign is initialized with the seed molecule, protein target,
optimization objectives, iteration limit, and batch size extracted from
the user request. A shared state records the enumerated candidate
library, previously evaluated molecules, returned characterization
values, and optimization history, allowing information generated at one
stage to be accessed by subsequent workflow nodes.

During each iteration, the Bayesian optimization node recommends a batch
of unevaluated candidates, which are routed to the characterization
tools associated with the specified objectives. The returned values are
appended to the campaign history and used to update the Gaussian-process
surrogate before the next batch is selected. Conditional routing
continues this cycle until the iteration limit is reached, the
predefined convergence criterion is satisfied, or the enumerated search
space is exhausted. LangGraph checkpointing preserves intermediate
workflow states, enabling interrupted campaigns to resume without
repeating completed characterization steps.

\subsection{Single-Objective Optimization}

\begin{figure}
	\centering
	\includegraphics[width=1\linewidth]{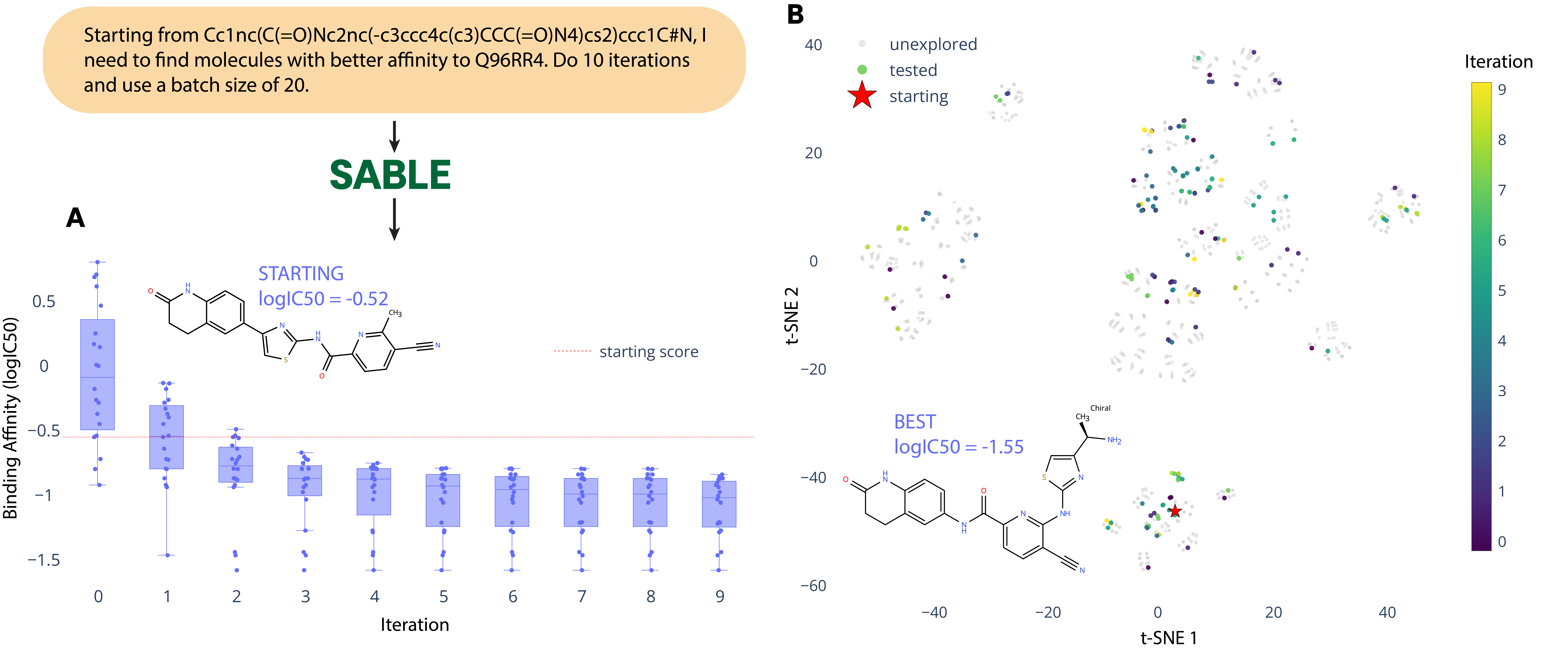}
	\caption{
Single-objective optimization of binding affinity.
Starting from a ligand specified by its SMILES string, SABLE enumerates
analogs and iteratively finds molecules with higher affinity for the
target with UniProt ID Q96RR4. (A) Box plots show the
cumulative distribution of the 20 lowest predicted Boltz-2 scores (in
logIC50). Lower values correspond to stronger predicted affinity. The
red dotted line indicates the seed compound\textquotesingle s logIC50 of
--0.52; the best-observed candidate had a predicted value of --1.55.
(B) t-SNE visualization of the enumerated analog space. Colored
points indicate evaluated candidates, with color denoting the iteration
in which each candidate was evaluated; grey points indicate candidates
that remained unevaluated at campaign termination. The structure and
predicted logIC50 of the best-observed candidate are shown.}
	\label{fig:fig2}
\end{figure}

We first evaluated whether SABLE could optimize a single computational
objective using human calcium/calmodulin-dependent protein kinase kinase
2 (CAMKK2; UniProt Q96RR4) as the
target\cite{kanevKLIFSOverhaulFirst2021,vanlindenKLIFSKnowledgeBasedStructural2014}.
Starting from a seed compound, HEALER generated a reaction-constrained
analog library, from which Bayesian optimization iteratively selected
candidates for Boltz-2
evaluation\cite{passaroBoltz2AccurateEfficient2025a} (Fig. \ref{fig:fig2}A).

Across the campaign, the best-observed predicted logIC50 improved by
1.03 log units relative to the seed, while the cumulative distribution
of the 20 best-observed candidates plateaued after approximately five
iterations (Fig. \ref{fig:fig2}A). Evaluated compounds were distributed across the
projected analog space, although most of the enumerated library remained
unevaluated at campaign termination (Fig. \ref{fig:fig2}B). Thus, SABLE prioritized
candidates with improved values of the assigned computational objective
while characterizing only a subset of the available library; this
behavior is characteristic of BO acquisition functions, which balance
exploration of uncertain regions with exploitation of promising
candidates.

\subsection{Dual-Objective Optimization}

We next evaluated whether SABLE could prioritize candidates across two
objectives by jointly optimizing Boltz-2-predicted affinity and a
quantitative estimate of drug-likeness
(QED)\cite{bickertonQuantifyingChemicalBeauty2012a}. Across successive
iterations, the evaluated candidates produced a set of nondominated
solutions spanning different compromises between lower predicted logIC50
and higher QED (Fig. \ref{figs:figs3}). Some candidates favored one objective more
strongly, whereas others achieved more balanced profiles (Fig. \ref{figs:figs3}). The
resulting Pareto frontier accordingly provides a set of alternative
candidates rather than a single optimum, allowing subsequent selection
to reflect project-specific priorities. These results show that SABLE
can extend beyond single-objective optimization to prioritize candidate
sets across an explicitly defined multi-objective landscape. All
campaigns beyond one objective are guided with the qLogNEHVI acquisition
function\cite{amentUnexpectedImprovementsExpected2025,daultonParallelBayesianOptimization2021},
to maximize the expected hypervolume improvement. This approach
evaluates candidate molecules based on how much they are expected to
expand the Pareto front relative to previously evaluated compounds.

\subsection{Multi-Objective Optimization}

\begin{figure}
	\centering
	\includegraphics[width=1\linewidth]{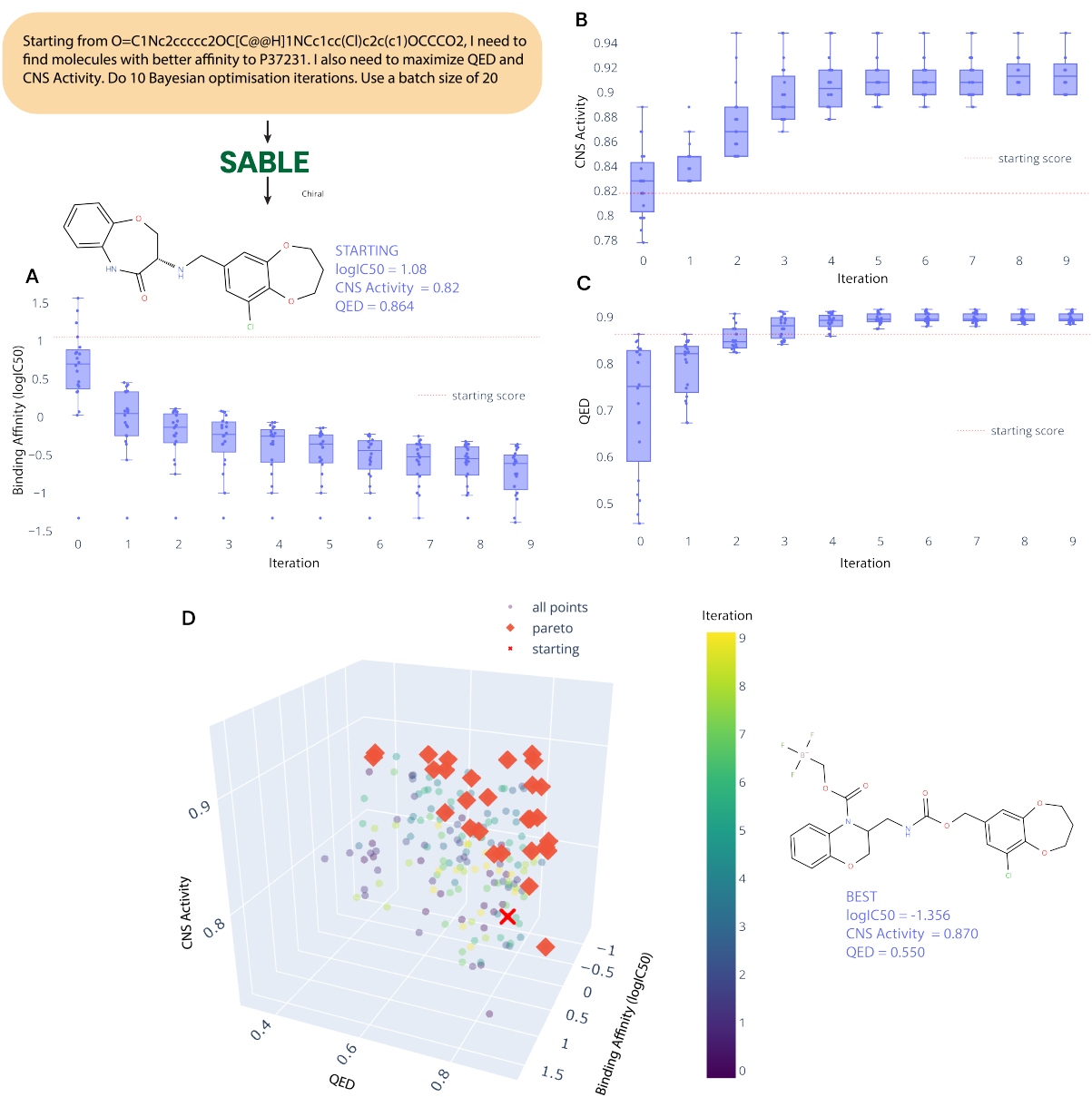}
	\caption{
The multi-objective optimization of affinity, QED, and
CNS activity. (A) Boxplots showing the affinity distribution in logIC50
of the best 20 ligands evaluated so far in the optimization campaign.
The structure of the starting compound is shown with logIC50, CNS
Activity, and QED scores of 1.08, 0.82, and 0.86, respectively. The red
dotted line indicates the score of the starting compound. (B)
Boxplots showing the distribution of the best 20 ligands that optimize
the CNS activity objective. (C) Boxplots showing the
distribution of the best 20 ligands that optimize the QED objective.
(D) 3D plot showing the Pareto frontier of all three
optimization objectives. The best compound resulting from the
optimization campaign is also pictured. We observe improvements in
binding and CNS activities, but a lower QED based on the tradeoff.}
	\label{fig:fig3}
\end{figure}

To further test the framework under more realistic drug discovery
constraints, we performed optimization experiments that simultaneously
considered multiple properties: Binding affinity, QED, and CNS Activity.
SABLE can optimize additional ADMET and physicochemical properties (SI
Note S5). Across multiple iterations, SABLE progressively sampled the
candidate library to identify molecules that occupy favorable regions of
the objective space (Fig \ref{fig:fig3}A - C). Early iterations emphasized
exploration to establish an initial Pareto frontier, while later
iterations concentrated on regions predicted to yield balanced
improvements across several properties (Fig. \ref{fig:fig3}D). The resulting enriched
libraries contained molecules that simultaneously satisfied multiple
drug-like criteria, illustrating the ability of SABLE to perform
automated multi-parameter optimization within a chemically constrained
search space. From the resulting Pareto set, we chose a single compound
that satisfies our tradeoff and present it as the `best'. In this work,
we prioritize higher affinity scores over other objectives.

\subsection{Efficiency of exploring the chemical space}

One of the major challenges in ligand discovery is the enormous size of
accessible chemical space. Even modest combinatorial modifications of
the scaffold can yield millions of potential analogs. Exhaustive
evaluation of these molecules is computationally infeasible,
particularly when structure-based affinity predictions are required. To
examine whether SABLE could effectively select optimal molecules from a
large sample, we benchmarked the optimization routine on
one-million-compound libraries using QED as the objective because it
could be calculated for every molecule in this library. While there is
criticism on using QED as a benchmark to compare across optimization
methods\cite{trippFreshLookNovo2021}, our analysis is appropriate to
study how the optimization method affects the time-to-best-compound
within a large
library\cite{gaoSampleEfficiencyMatters2022,korovinaChemBOBayesianOptimization2020}.Separate
campaigns were conducted using libraries generated by similarity- and
diversity-based sampling (Fig. \ref{figs:figs4}). We found that the optimizer
recovered the global maximum after three iterations in the
similarity-sampled library and seven iterations in the diversity-sampled
library. The efficiency gain becomes even more pronounced as the number
of objectives increases, since multi-objective optimization rapidly
becomes intractable under exhaustive screening. Quantitatively, across
the three campaign classes (sections 2.2 - 2.4), SABLE evaluated only a
fraction of the enumerated library before convergence, translating to a
reduction in Boltz-2 inference calls relative to exhaustive screening.
The relative savings in oracle or experimentation calls, grow
approximately linearly with library size in the regime of ⟨10\textsuperscript{3} to
10\textsuperscript{6}).

\subsection{Case study 1: Retrospective METTL3 lead optimization}

We next evaluated SABLE in a retrospective medicinal chemistry setting
using the METTL3 inhibitor campaign reported by Dolbois et al. \cite{dolbois149Triazaspiro55undecan2oneDerivativesPotent2021a}
This case study is useful since the published series contains a clearly
defined experimental trajectory that optimizes an early METTL3 hit into
the potent and selective inhibitor UZH2, with a reported single-digit
nanomolar biochemical IC50 and favorable cellular target engagement. We
supplied SABLE with the starting compound from the paper, specified
METTL3 as the target using UniProt ID Q86U44, and asked the system to
minimize predicted binding affinity using our agentic workflow.

The starting compound was predicted by Boltz-2 to have an IC50 of 9.77
~$\mu\mathrm{M}$, which is consistent with the reported low-micromolar experimental
activity of the starting compound
(7~$\mu\mathrm{M}$)\cite{dolbois149Triazaspiro55undecan2oneDerivativesPotent2021a}.
For the published optimized compound UZH2, Boltz-2 predicted an IC50 of
31.6 nM, whereas the experimental TR-FRET IC50 was 5nM. Thus, Boltz-2
moves in the direction of increased predicted affinity. The distinction
is important since Boltz-2 in SABLE only serves as a computational
oracle for enrichment and not a substitute for measured biochemical
potency. However, the results of this exercise illustrate that SABLE
will assure success in cases when computed ligand scores correlate with
experimental measurements.

\begin{figure}
	\centering
	\includegraphics[width=1\linewidth]{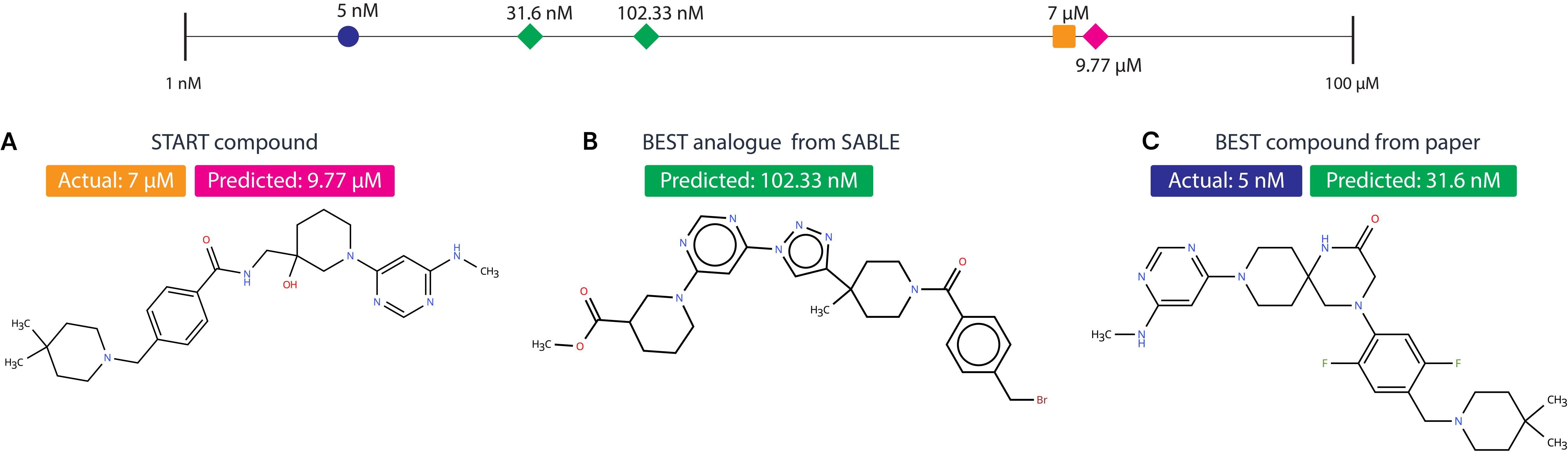}
	\caption{
Retrospective METTL3 lead-optimization case study.
(A) The starting compound is shown with both the reported experimental
IC50 and the Boltz-2-predicted concentration. (B) The best SABLE analog
is shown as a Boltz-2-predicted value since it has not been
experimentally tested. (C) The published best compound is shown with
both its experimental IC50 and Boltz-2-predicted concentration.}
    \label{fig:fig4}
\end{figure}

Despite this limitation, the retrospective run illustrates the practical
value of the agent. Without using any experimental feedback from the
METTL3 campaign, SABLE identified a top-ranked analog with a predicted
concentration of 102.33 nM (Fig. \ref{fig:fig4}). Relative to the starting compound,
this represents approximately a 95-fold decrease in predicted
concentration from a single optimization run. The candidate was not
experimentally tested, so the result is interpreted as a prospective
prioritization rather than experimental validation. Nevertheless, it
demonstrates the ability of SABLE to move from a low-micromolar starting
point to a predicted near-100 nM candidate for a challenging target such
as METTL3.

The comparison also highlights a limitation of the current
implementation. SABLE did not rediscover the exact published lead
compound. The HEALER library used in this run did not include the
reaction template required to construct that final
triazaspiro{[}5.5{]}undecan-2-one series. The search space definition,
therefore, constrained which structural modifications were available to
the optimizer. The system can efficiently prioritize molecules within
the accessible chemistry it is given, but its performance is limited by
the coverage of the enumeration engine and the calibration of the
characterization model. Expanding the reaction-template set and
incorporating SABLE into experimental DMTA campaigns are therefore
expected to improve both the chemical reach and the reliability of
future campaigns.

\subsection{Validation case study 2: Optimization on Boltz-2 tested
targets} 
Because SABLE currently uses Boltz-2 as its structure-based
affinity oracle, a natural test is to ask whether SABLE can improve
ligands on targets for which Boltz-2 itself reports target-level
affinity validation. Notably, the universal performance of Boltz-2
across all targets has been questioned \cite{bretAssessingBoltz2Performance2026, wanReliabilityAIMethods2026}, which is why we sought
targets where Boltz-2 performed well to examine the effectiveness of
SABLE in the absence of potential failure attributable to the scoring
function. In this case study, we select a few targets that Boltz-2 can
rank with useful signals and use SABLE to turn a weak compound from an
experimental series into analogs that have the same oracle scores as
competitive with, or better than, the best in that series.

We selected four targets from the Boltz-2 affinity-validation set that
span distinct target classes: beta-secretase 1 (P56817, protease),
carbonic anhydrase XII (O43570, lyase), ABL1 (P00519, kinase), and
sphingosine-1-phosphate receptor 1 (P21453, GPCR). These targets were
useful for this case study since Boltz-2 reports per-assay validation
performance for each of them in its hit-to-lead affinity benchmark, with
positive rank and linear correlations across all four targets \cite{passaroBoltz2AccurateEfficient2025a}.
We then used the ChEMBL assay data for each target to define a
deliberately difficult starting point. For each assay, the
censored-low-activity compound in the experimental series was used as
the seed molecule, while the best ChEMBL compound in the same assay
provided a reference point for what a strong ligand in that series
looked like. The ChEMBL data was used only for seed and reference
selection. SABLE was not given the SAR trajectory or experimental
activity labels during optimization; the only optimization signal was
the Boltz-2 affinity score returned for candidates selected from the
enumerated analog library.

\begin{figure}
	\centering
	\includegraphics[width=1\linewidth]{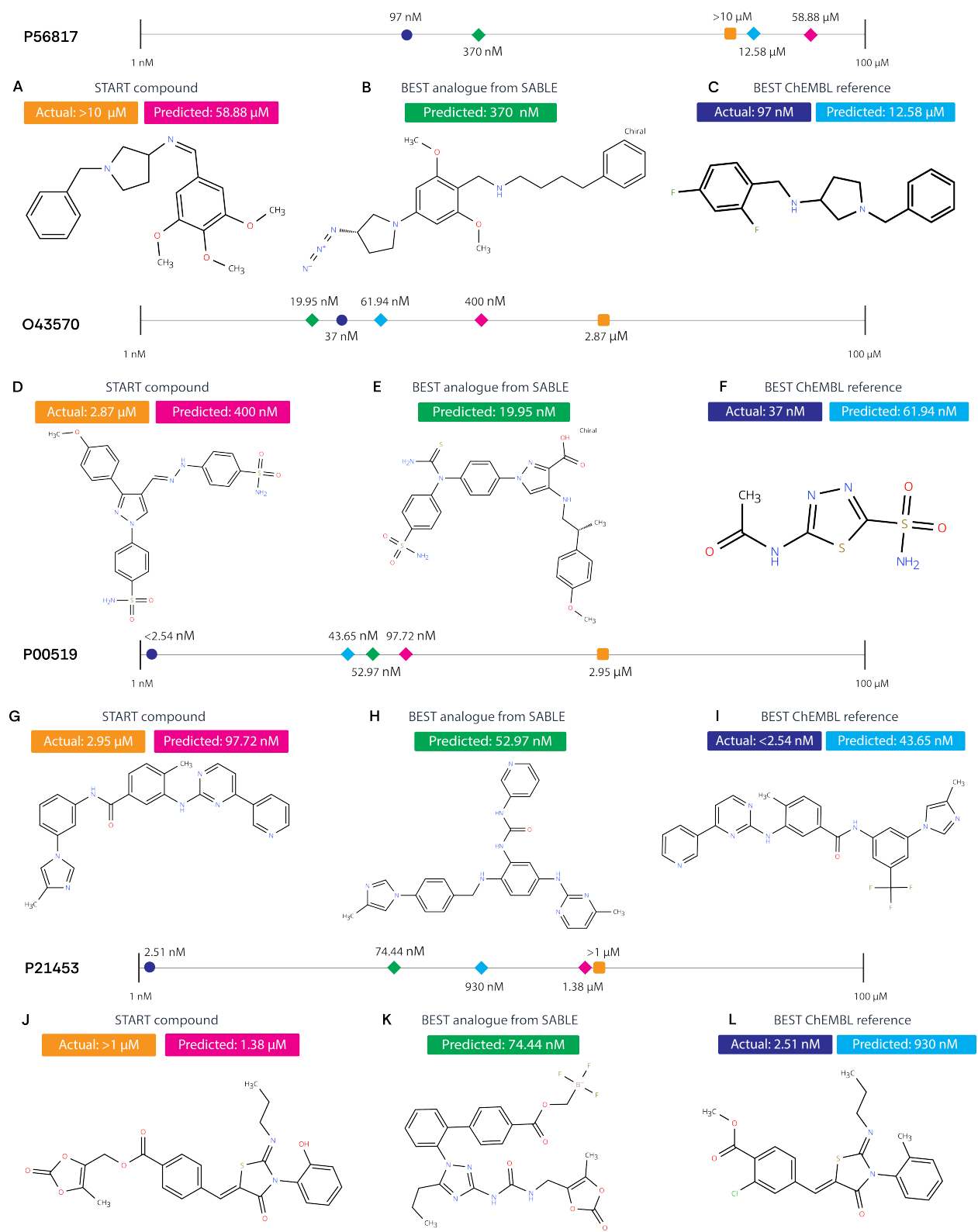}
	\caption{
Ligand and affinity comparison for the retrospective
Boltz-2 target validation. Each panel shows the ChEMBL seed compound
selected from the weak end of the assay series, the best ChEMBL
reference compound from the same series, and the best analog identified
by SABLE. Lower concentration values indicate stronger activity or
predicted affinity.}
    \label{fig:fig5}
\end{figure}

We report this case study directly in the original concentration units
(Table \ref{tab:table1} and Fig. \ref{fig:fig5}). The ChEMBL columns show experimental IC50 values
from the assay series, while the Boltz columns show predicted
concentrations from the SABLE runs. Lower concentrations indicate
stronger activity or predicted affinity. For censored ChEMBL records,
the original bound is retained. Across all four targets, SABLE
identified analogs with improved Boltz-predicted affinity relative to
the starting compound. The largest predicted shift was observed for
P56817, where the seed was predicted at 58.88 ~$\mu\mathrm{M}$ and the best SABLE
analog at 370 nM. O43570 and P21453 also showed substantial predicted
shifts, from 400 nM to 19.95 nM and from 1.38 ~$\mu\mathrm{M}$ to 74.44 nM,
respectively. P00519 began from a seed that Boltz already scored as
considerably stronger than its ChEMBL assay value, but SABLE still
improved the predicted value from 97.72 nM to 52.97 nM.

\begin{table}
	\caption{SABLE optimization on Boltz-2 affinity-validation
targets, reported in experimental and predicted concentration units}
	\centering
	\small
	\setlength{\tabcolsep}{3pt}
	\renewcommand{\arraystretch}{1.15}
	\begin{tabularx}{\textwidth}{@{}l>{\raggedright\arraybackslash}X*{5}{>{\raggedright\arraybackslash}p{0.105\textwidth}}@{}}
		\toprule
		Target & ChEMBL assay reference & 
    ChEMBL value for seed molecule & Boltz-2 prediction for seed & 
    Best ChEMBL reference & Boltz-2 prediction for ChEMBL best & Best from SABLE \\
		\midrule
P56817 & Beta-secretase 1 / BACE-1 FRET assay & \textgreater{} 10~$\mu\mathrm{M}$ &
58.88~$\mu\mathrm{M}$ & 97 nM & 12.58~$\mu\mathrm{M}$ & 370 nM \\
O43570 & Carbonic anhydrase XII / spectrophotometric & 2.87~$\mu\mathrm{M}$ & 400 nM
& 37 nM & 61.94 nM & 19.95 nM \\
P00519 & ABL1 / peptide-substrate inhibition assay & 2.95~$\mu\mathrm{M}$ & 97.72 nM
& \textless{} 2.54 nM & 43.65 nM & 52.97 nM \\
P21453 & S1PR1 / beta-arrestin recruitment assay & \textgreater{} 1~$\mu\mathrm{M}$ &
1.38~$\mu\mathrm{M}$ & 2.52 nM & 930 nM & 74.44 nM \\
		\bottomrule
	\end{tabularx}
	\label{tab:table1}
\end{table}

The case study shows that SABLE can use Boltz-2 as an optimization
oracle in the setting for which Boltz-2 was designed, that is, in
ranking compounds within hit-to-lead chemical series. While SABLE relies
on Boltz-2, which does not provide excellent correlation between
measured and predicted affinities of top binders, it still converges on
molecules that score more than two orders of magnitude better than the
starting points. Starting from the experimentally weakest compounds in
the ChEMBL assay series, SABLE enriched the local analog space for the
molecules with substantially lower predicted concentration values from
their seed compounds. The final predicted values are also in the same
broad range as the experimental activities of the best ChEMBL reference
compounds for P56817, O43570, and P21453. For P00519, the seed
prediction was predicted to be much more active, exposing a limitation
in using Boltz-2 as an oracle.

SABLE enriches compounds against the Boltz-2 objective, and not against
unmeasured biochemical truth. The result therefore supports the intended
role of SABLE as a prioritization engine for proposing analogs that
should be considered for synthesis or secondary computational analysis,
rather than as a replacement for experimental potency measurements. In
an experimental closed-loop deployment, the same workflow could replace
or augment the Boltz-2 characterization node with real assay results as
they become available, allowing the optimizer to use real-life
experimental data while retaining the same agentic orchestration and
selection framework.

\section{Discussion}

A key architectural design in SABLE is the distribution of the
respective contributions between LLMs, QSAR, and optimization tools. The
LLM does not optimize molecules but translates human intent into
machine-readable optimization configurations; resolving molecule names
to canonical SMILES, preparing node-routing configurations, and
surfacing results back to the user. It is never used to propose analogs,
score affinity, or reason about ADMET liabilities. Each of those
decisions is delegated to a purpose-built component with validated
statistical or physical grounding: HEALER for synthetically constrained
enumeration, Bayesian Backend (BayBE) for BO, Boltz-2 for
structure-based affinity, and dedicated QSAR predictors for ADMET
endpoints. This responsibility split is what distinguishes SABLE from
other systems like ChemCrow or Coscientist that place the LLM at the
center of scientific reasoning. This design is informed by evidence that
LLMs can hallucinate chemically invalid structures, misreport
physicochemical properties, and are biased toward molecular patterns
overrepresented in their training corpus. A frequent critique of
tool-augmented agents like SABLE is that they trade fluency of an
end-to-end LLM for engineering complexity
\cite{kujanpaaToolMakingSelfEvolvingLLM2026}. Our results show that this
tradeoff is justified in a technical domain such as drug discovery.
Across single- and multi-objective campaigns (Figs. \ref{fig:fig2}-\ref{fig:fig3}), SABLE produced
improvements over the seed compound on every tracked objective, and the
resulting enriched compounds consisted entirely of molecules
constructible from commercially available building blocks under known
reaction templates. This means that the resulting compounds are directly
actionable by medicinal chemistry teams, in contrast to generative
outputs that must first be triaged for feasibility. The BO layer further
means that useful enrichment is obtained after evaluating only a small
fraction of the enumerated space (Fig. \ref{fig:fig2}), and efficiency gains
compounds as the number of objectives grows and exhaustive evaluation
becomes combinatorially infeasible.

The computational chemistry tool development landscape moves quickly,
making it impractical to hardcode a particular affinity predictor or
QSAR model should be replaced as soon as a better model or tool become
available. SABLE is therefore designed around a standardized
tool-integration interface that treats characterization and enumeration
backends as both interchangeable and extendable. Swapping Boltz-2 for
another affinity model or augmenting the ADMET stack with newly
published toxicity predictors requires no change to the orchestration
graphs or optimization logic, only a node-level adapter. The same
principle extends to the enumeration layer, where HEALER could be
replaced or complemented by other tools with similar functionality such
as SynFormer or reaction-based generative models as they mature. This
modularity positions SABLE as a non-static workflow, a substrate on
which the community can build successively stronger optimization
pipelines (Fig \ref{figs:figs2}). In addition, SABLE's architecture makes it possible
to wrap the graph around laboratory equipment via REST APIs or Model
Context Protocol (MCP) endpoints, allowing characterization nodes to
dispatch real synthesis and assay requests rather than purely
computational predictions. Because SABLE's optimization loop is already
framed around the DMTA paradigm and treats each characterization as a
black-box oracle, the transition from simulated to experimental
measurements is primarily an engineering integration rather than an
algorithmic change. This design choice helps bridge the computational
hit-to-lead optimization with closed-loop autonomous laboratories as the
underlying robotic infrastructure matures.

Several caveats currently limit the agentic workflow. First, every claim
SABLE makes about a candidate molecule is downstream of a predictive
model, and predictive models for affinity, permeability, and toxicity
remain imperfect; a library that is "enriched" in Boltz-2 logIC50 is
enriched against its limitations, not against experimental ground truth.
This motivates the eventual coupling to laboratory automation, where
real measurements can close the loop. Second, HEALER's synthetic
accessibility is a necessary but insufficient condition for true
experimental tractability. HEALER, by design, provides existing building
blocks along with a reaction template, but this information alone does
not constitute an experimental protocol. Furthermore, HEALER is designed
to produce synthetically similar compounds; it does not provide the
opportunity to assess structurally similar but synthetically
dissimilar compounds; in the future, we will add the capability to
explore this important category of chemical modifications that do not
consider synthetic closeness.

Third, the Gaussian-process surrogate scales polynomially with each
evaluated compound, and while this is tractable for conservative batch
sizes, very large search spaces and long campaigns will eventually
require space or deep-kernel approximations.

\section{Conclusions}

SABLE is an open-source agentic platform for ligand optimization. It can
act in the lead-optimization part of the drug discovery pipeline, and it
does this by executing a computational analog of the DMTA cycle that
mirrors modern medicinal chemistry campaigns. Across single and
multi-objective experiments, SABLE consistently enriched libraries for
predicted property, drug likeness, and other relevant physicochemical
properties, producing frontiers of synthesizable molecules that improved
on a user-provided seed compound while evaluating only a small fraction
of the enumerated search space.

Two properties distinguish this workflow. First, every molecule SABLE
proposes is makeable from commercially available building blocks through
retrosynthetic reaction templates, which eliminates failures that arise
when generative models yield synthetically infeasible molecules. Second,
the agent is designed for extension: its tool-integration interface
allows any component like an affinity predictor, an ADMET stack, or an
enumerator to be swapped for newer and better tools without touching the
orchestration logic. Together, these properties position SABLE as a
durable substrate for hit-to-lead optimization rather than a snapshot of
today's best tools.

\section{Methods}

\subsection{Argument Extraction} Every task on SABLE begins with an
argument extraction after a prompt is sent. The argument extraction node
employs a hybrid extraction mechanism that combines rule-based regex
parsing with LLM-assisted interpretation. The LLM component can handle
ambiguous or complex prompts, resolving molecule names to structures and
inferring implicit optimization objectives. From a prompt, SABLE can
extract the starting molecule (SMILES), protein target (UniProt ID),
optimization target (binding affinity, minimize), enumeration size,
iteration count, and batch size. Importantly, SABLE automatically
detects the number and types of objectives present in the user prompt
during the argument extraction state (Table \ref{tabs:tables2}). These objectives are
passed to the Bayesian optimization node, which dynamically configures
the optimization strategy to handle either single-objective or
multi-objective problems without requiring user intervention. All prompt
extractions were done with GPT-5-chat preview (version 2025-10-03),
provided by Azure OpenAI.

\subsection{Name to Entity} After argument extraction, SABLE runs a
"name-to-entity" step to normalize user-specified molecular and protein
identifiers into machine-readable structures. This module maps common
compound/drug names (e.g., aspirin, ibuprofen) to SMILES strings,
enabling downstream generation and scoring without requiring users to
provide structures explicitly. For target proteins, it accepts UniProt
IDs and retrieves the corresponding structural entities used by the
optimization workflow.

To improve robustness, the node uses a hybrid resolver that combines
deterministic identifier matching with LLM-assisted disambiguation for
incomplete or ambiguous names. Retrieved molecules are standardized
(e.g., canonicalized SMILES) before being passed to later stages, while
protein entities are validated against available structure metadata.
This ensures that both ligand and target inputs are consistently
represented before Bayesian optimization begins.

\subsection{Molecular search space enumeration} Ligand generation
carries the requirements that the ligands are both synthesizable and
rationally permutable from a starting compound -- without these two
features, evolution of ligands toward more optimal candidates cannot
proceed. We incorporated HEALER as a molecular enumeration
tool\cite{kelestemurHEALERHitExpansion}. HEALER workflow contains three
modes of molecule enumeration that serve slightly different purposes.
Molecule-mode is the default way of generating synthesizable analogs of
a given seed molecule. HEALER initially performs a recursive
retrosynthetic fragmentation of the query molecule using predefined
reaction templates to create a retrosynthesis tree with fragment and
reaction annotations. In the next step, HEALER matches building blocks
to fragments based on chemical similarity and recombines them via
predefined reactions to produce analogs of the seed molecule. Similarly,
Fragment-mode allows the enumeration to directly start from fragments
and generate molecules that combine the given fragments in a
synthetically feasible way. Lastly, site-mode allows the reaction
enumerations to take place at specified reactive sites of the seed
molecule while preserving the rest. In this study, Enamine US building
block stock was used by HEALER, and Molecule-mode was allowed to have at
most two retrosynthesis reaction steps to prevent long enumeration
runtimes.

This approach addresses the molecular hallucination problem. Unlike
generative models that propose arbitrary SMILES strings, every molecule
in SABLE's search space is constructed from these purchasable fragments.
HEALER also accepts limits on the chemical similarity of evolved
compounds, providing a tunable parameter to balance between fine-grained
and coarse-grained exploration of the chemical space. Given the
modularity of SABLE, the enumeration node is tool-agnostic; that is, any
other molecular enumeration tool can be used in place of HEALER, such as
SynFormer\cite{gaoGenerativeAINavigating2025}.

\subsection{Optimization Node} Most agentic platforms for chemical
discovery use LLMs to iteratively suggest chemical perturbations to
improve the desired properties of a discovery campaign
\cite{averlyLIDDIALanguagebasedIntelligent2025a,liuDrugAgentAutomatingAIaided2025,m.branAugmentingLargeLanguage2024a}.
This is done under the assumption that LLMs can apply chemical
rationale. While this can work for short-horizon and single-objective
optimization, it is subject to significant bias arising from LLM
training and may not adequately explore the chemical space when the task
allows for broad exploration and sensible chemical modifications.
Furthermore, multi-objective optimization is not tractable with this
approach. We integrated an optimization node that allows SABLE to
perform both single- and multiple-objective optimization of molecular
properties. Bayesian optimization is used in this case; it is a
sequential design method for globally optimizing black-box functions. In
the case of lead optimization, a black-box function has no interpretable
functional form and is hard to measure. Examples of these are binding
affinity, toxicity, etc. Our Bayesian optimization module consists of a
Gaussian process surrogate model and a
\emph{qLogExpectedImprovement(qLogEI)} acquisition function. In the case
of multi-objective optimization, we utilize the
\emph{qLogNoisyExpectedHypervolumeImprovement (qLogNEHVI)} acquisition
function.

The optimization node uses BayBE's Campaign API to recommend the next
batch of molecules for evaluation. Molecules are encoded as
\emph{SubstanceParamater} objects using one of several fingerprint
schemes: Mordred, Extended Connectivity Fingerprints (ECFP), or RDKit
descriptors. The surrogate model is updated with all prior experimental
results at each iteration, and the acquisition function selects the most
promising unevaluated molecules. The system validates recommendations
against the search space and filters out previously tested molecules
before forwarding them for characterization. Extended information about
this node is provided in (SI Notes S3 -- S5).

\subsection{Characterization Nodes} SABLE integrates characterization
backends that are selected automatically based on the optimization
target of interest. \textbf{RDKit} computes molecular descriptors such
as QED, LogP, TPSA, molecular weight, H-bond donors/acceptors, rotatable
bonds, ring count, etc \cite{RDKit}. \textbf{STOPLIGHT} predicts
physicochemical and structural properties, assay liabilities, and
pharmacokinetic properties for small molecules
\cite{wellnitzSTOPLIGHTHitScoring2024b}. \textbf{Boltz-2} accepts ligand
SMILES and protein sequences (or UniProt IDs, resolved automatically)
and returns affinity estimates. A \emph{`decide\_characterization'} node
in the agent's graph inspects the target properties and selects the
minimal set of tools required. If all targets are RDKit-computable, only
RDKit is invoked. Binding affinity targets trigger Boltz-2. Properties
spanning both toolsets invoke a combined pipeline.

\subsection{Binding Affinity Scoring} Binding affinity prediction was
performed using Boltz-2\cite{passaroBoltz2AccurateEfficient2025a}, a
deep learning model for biomolecular structure and property prediction
based on a diffusion-transformer architecture. Candidate molecules
selected by the optimization module are provided to the model as SMILES
strings, while the amino acid of the target protein is retrieved by the
argument extraction tool. The Boltz-2 YAML input is prepared from the
provided SMILES and target sequence information. Boltz-2 jobs are
dispatched asynchronously to a dedicated inference backend. In API mode,
the YAML payload is submitted via an authenticated HTTP request to a
REST endpoint, and the client polls for jobs at intervals; all ligands
within a given optimization batch are processed concurrently using
asynchronous I/O. Alternatively, in Celery mode, jobs are submitted to a
SLURM-managed high-performance computing cluster via a task queue, with
execution performed inside an Apptainer container, enabling
GPU-accelerated inference. Upon completion, the model returns an
affinity prediction value extracted from the per-job JSON output. This
value is recorded and passed back to the optimization module as the
objective signal for subsequent molecule selection. The predicted 3D
structure of the protein-ligand complex is additionally retrieved and
stored as a CIF file, providing structural interpretability alongside
the scalar affinity score.

\subsection{Stopping Criteria} Optimization loops within SABLE
terminate when one of several predefined conditions is met. These
stopping criteria prevent unnecessary computation once the search
process has converged or the search space has been exhausted. (i)
Maximum number of iterations specified by the user has been reached.
This parameter provides an upper bound on runtime and computational
cost. (ii) The convergence node monitors improvement in the objective
function across iterations. If no significant improvement is observed
over several consecutive iterations, the system interprets this as
convergence of the optimization process. (iii) The workflow terminates
if all molecules within the enumerated search space have been evaluated.
This condition typically occurs only when the search space is relatively
small. (iv) Optimization stops if the remaining unevaluated molecules
are fewer than the specified batch size. In this case, the system
evaluates the remaining candidates and completes the campaign.

\subsection{Webapp and user interface} SABLE is distributed with a CLI,
but it also includes a web-based interface designed to simplify
interaction with the optimization workflow. The interface allows users
to submit prompts describing optimization objectives, seed molecules,
and monitor the progress of active optimization campaigns. The backend
server manages agent execution and dispatches tasks to the appropriate
computational modules. Results from each iteration are stored in a
database and streamed to the interface, allowing users to inspect
intermediate results such as predicted affinities, molecular properties,
and candidate structures. The visualization components allow the results
to be inspected and downloaded. The full architecture is shown in Fig.
\ref{figs:figs2}.

\subsection{Handling ambiguous questions} User prompts may contain
ambiguous or incomplete information regarding optimization targets,
molecule identity, or experimental constraints. If the prompt lacks
certain parameters, the system uses default configurations. This process
ensures that optimization campaigns begin with well-defined parameters
while still allowing for cases where the user might not be familiar with
optimization jargon. When critical information is missing, such as the
starting molecule name or a protein target required for affinity
prediction, the system fails early and informs the user of what they
might need to provide.

\subsection{Computational environment and configurations} All SABLE
experiments reported in this text were performed in a Dockerized
environment on a single laptop. The Docker containers ran on Ubuntu
22.04.5 LTS via the Windows Subsystem for Linux (WSL), restricted to 10
CPU cores and 32 GB of RAM. Binding affinity estimations with Boltz-2
were carried out on a compute node equipped with four NVIDIA L40 GPUs
(46 GB each). Each protein--ligand affinity inference was carried out in
an Apptainer (Singularity) environment with a single GPU.

\section{Code and Data Availability}

SABLE is released under an Open Source Apache license and can be found
at \url{https://github.com/molecularmodelinglab/SABLE}\textbf{.} The
Dockerfiles and Apptainer SIF reproducing the exact software stack are
released alongside the code. The prompts and JSON state-dumps required
to reproduce every figure are also deposited in the repository.

\section{Acknowledgements}

This research was supported by funding from NSF grant 2324394 and NIH
grant GM140154. Compute resources for this work were provided by the 
AI Acceleration Program at the University of North Carolina at Chapel Hill.

\section{Author Contributions}

\textbf{KPI}: Conceptualization, Methodology, Software, Formal
analysis, Investigation, Data Curation, Visualization, Writing -
Original Draft; \textbf{EK}: Conceptualization, Methodology, Software,
Data Curation, Writing - Original Draft; \textbf{BS}: Conceptualization
and Writing - Original Draft; MH: Methodology and Software; \textbf{SD:}
Methodology; \textbf{AA:} Methodology; \textbf{RA:} Methodology,
Supervision; \textbf{MD}: Conceptualization, Project administration,
Supervision, Writing -- Review \& Editing; \textbf{AT}:
Conceptualization, Supervision, Writing -- Final Review \& Editing.

\section{COI}

AT is a co-founder of Predictive LLC, a company specializing in
predicting toxicological profiles of small molecules for drug discovery.
All other authors have no conflicts to declare.

\printbibliography

\pagebreak
\clearpage
\begin{center}
\textbf{\large Supplemental Information}
\end{center}

\setcounter{equation}{0}
\setcounter{figure}{0}
\setcounter{table}{0}
\setcounter{section}{0}
\makeatletter
\begin{refsection}
\begin{refcontext}[labelprefix=S]
\renewcommand{\thefigure}{S\arabic{figure}}
\renewcommand{\thetable}{S\arabic{table}}
\renewcommand{\thesection}{S.\arabic{section}}

\section{Molecule Encoding}

Molecules entering the optimization campaign are wrapped as
\emph{SubstanceParameter} objects, a discrete label-like parameter type
that maps human-readable molecule labels to canonicalized SMILES
strings. Encoding from chemical structure to a numerical feature vector
consumable by the Gaussian process surrogate proceeds in four stages.

\textbf{Stage 1:} \textbf{SMILES validation and canonicalization}. Each
SMILES string supplied is canonicalized via RDKit, and duplicates are
rejected. This guarantees a one-to-one correspondence between labels in
the search space and points in the descriptor space, preventing
degenerate rows in the kernel Gram matrix that would otherwise
destabilize the posterior.

\textbf{Stage 2: Fingerprint/descriptor computation.} SABLE can use
three encoding schemes through the BayBE \emph{SubstanceEncoding} enum,
selectable per campaign.

\begin{itemize}
\item
  MORDRED (default): a \textgreater1,600-dimensional vector of 2D
  molecular descriptors covering topological, constitutional, geometric,
  and electronic properties.
\item
  ECFP (Extended-Connectivity Fingerprints)\textbf{:} radius-3 circular
  fingerprints folded into a 2048 length bit vector, capturing local
  atomic environments.
\item
  RDKIT\textbf{:} the native RDKit descriptor set (\textasciitilde200
  descriptors), spanning physicochemical properties such as logP, TPSA,
  and ring counts.
\end{itemize}

\textbf{Stage 3: Cleaning.} Once the raw descriptor matrix is computed,
columns containing any \emph{NaN} values and columns with zero variance
are removed. These columns carry no information and would either crash
the kernel computation or unnecessarily inflate the dimensionality.

\textbf{Stage 4: Decorrelation.} High-dimensional cheminformatics
descriptors are heavily redundant. We therefore apply a correlation
threshold filter, i.e., pairs of descriptors with correlation exceeding
0.7 are reduced to a single representative. As noted in the Bayesian
Backend (BayBE) documentation, for libraries on the order of tens of
thousands of molecules, this typically reduces descriptors to only
informative features. This step is necessary for the Gaussian process
since redundant features inflate the effective length-scale prior, slow
the marginal-likelihood optimization, and degrade out-of-sample
predictive variance. Finally, a small Gaussian perturbation is applied
to any rows that remain numerically identical after decorrelation,
ensuring the kernel matrix is strictly positive-definite.

The cleaned, decorrelated dataframe is cached and passed downstream as
the feature representation \(X \in \mathbb{R}^{d}\) for each candidate
molecule.

\section{Gaussian Process Surrogate}

The surrogate is a SingleTask gaussian process wrapped by BayBE's
\emph{GaussianProcessSurrogate} class. Given training data
\(\mathcal{D} = \{(x_i, y_i)\}_{i=1}^{n}\), the model defines a prior

\[f(x) \sim \mathcal{GP}(m(x), \kappa(x, x')),\]

with constant function \(m(x) = c\) and covariance kernel
\(\kappa(x,\ x')\) produced by the \emph{DefaultKernelFactory.} The
default factory adapts the kernel to the dimensionality and structure of
the search space, defaulting to a Matérn 5/2 kernel with automatic
relevance determination (ARD), one length-scale per input dimension:

\[K_{\nu}(x, x') = \frac{1}{\Gamma(\nu)2^{\nu - 1}}
\left( \frac{\sqrt{2\nu}}{\ell}d(x, x') \right)^{\nu}
K_{\nu}\left( \frac{\sqrt{2\nu}}{\ell}d(x, x') \right)\]

where \(d(x, x') = \lVert x - x' \rVert\) is the Euclidean distance,
\(K_{\nu}\) is the modified Bessel function of the second kind, and
\(\ell \in \mathbb{R}^{d}\) is the per-dimension
length-scale vector inferred by maximum marginal likelihood. The Matérn
5/2 kernel produces sample paths that are twice mean-square
differentiable, a smoothness assumption well matched to molecular
property surfaces, which are continuous functions of descriptor space
but typically not infinitely differentiable.

\textbf{Likelihood.} Observations are modeled as

\[y_i = f(x_i) + \varepsilon_i, \qquad
\varepsilon_i \sim \mathcal{N}(0, \sigma_n^2),\]

with a \emph{GaussianLikelihood.} The noise variance \(\sigma_{n}^{2}\)
is initialized and constrained by a data-driven prior produced, which
sets a weakly informed LogNormal prior over \(\sigma_{n}^{2}\) scaled to
the spread of the target variables. This prior regularizes the noise
estimate when n is small and prevents the marginal-likelihood optimizer
from collapsing \(\sigma_{n}^{2}\  \rightarrow \ 0\)

\section{Acquisition Functions}

SABLE uses logarithmic, batch (q-) Monte Carlo variants of expected
improvement throughout. These are provided through the BayBE library.
The log transformation reformulates expected-improvement acquisition
functions to avoid numerical underflow in regions where the analytics
improvement is small, which frequently occurs in late-stage lead
optimization once the surrogate is
well-trained\cite{amentUnexpectedImprovementsExpected2025}. This
improves the optimization of the acquisition surface relative to the
standard qEI

\subsection{Single-objective: qLogExpectedImprovement
(qLogEI)}

For a batch \(X = \{ x_{1},\ \ldots x_{q}\}\) and observed best value
\(f\left( X^{+} \right),\) the standard batch expected improvement is

\[qEI(X) = \ E_{Y\sim p\left( Y \middle| X,\ D \right)}\left\lbrack \max\left( 0,\ \max_{i}y_{i} - \ f\left( X^{+} \right) \right) \right\rbrack,\]

estimated by Monte Carlo with N samples drawn from the joint Gaussian
process posterior:

\[qEI(X)\  \approx \ \frac{1}{N}\sum_{j = 1}^{N}{\max\left( 0,\ \max_{i}y_{i}^{(j)}\  - \ f\left( X^{+} \right) \right).}\]

qLogEI replaces the non-smooth \(max(0,\ \ .)\) and the final
\(\max_{i}\) operators with their LogSumExp-based smooth analogs,
computing log-EI directly in a numerically stable way:

\[qLogEI(X) \approx {logsumexp}_{j}\left\lbrack {softplus}_{\tau}\left( \max_{i}^{\tau}\ y_{i}^{(j)} - f\left( X^{+} \right) \right) \right\rbrack - logN,\]

where \({softplus}_{\tau}(z) = \ \tau^{- 1}log(1 + exp(\tau z)\) and
\(\max^{\tau}\) is the temperature-\(\tau\) smooth maximum. Critically,
the gradient \(\nabla_{X}\ qLogEI(X)\) remains well-conditioned even
when EI itself has saturated to machine epsilon, allowing the
multi-start L-BFGS-B optimizer over the acquisition to reliably locate
batch optima that would be missed by qEI.

\subsection{Multi-objective: qLogNoisyExpectedHypervolumeImprovement
(qLogNEHVI)}

For multi-objective optimization, SABLE uses qLogNEHVI. Given M
objectives and a Pareto front \(\mathcal{P}_{\mathcal{D}}\) inferred
from the noisy training data, define the dominated hypervolume relative
to a reference point r as

\[HV\left( \mathcal{P;}r \right) = \ \lambda_{M}\left( \bigcup_{p\ \epsilon\ \mathcal{P}}^{}\lbrack r,p\rbrack \right),\]

where\(\lambda_{M}\) is the M-dimensional Lebesgue measure. The noisy
expected hypervolume improvement (qNEHVI) integrates the hypervolume
gain over both the posterior at the candidate batch \emph{and} the
posterior at the existing observations, marginalizing over noise on the
inferred front:

\[qNEHVI(X) = \ \mathbb{E}_{\widetilde{\mathcal{P}}\sim p\left( . \middle| \mathcal{D} \right)}\mathbb{E}_{Y\sim p\left( Y \middle| X,\mathcal{D} \right)}\left\lbrack HV(\widetilde{\mathcal{P}} \cup Y;r)\  - \ HV(\widetilde{\mathcal{P}};r) \right\rbrack.\]

qLogNEHVI applies the same LogSumExp smoothing as qLogEI to the
hypervolume difference, again yielding smooth, non-vanishing gradients
across the objective space and across the Pareto front. The reference
point r is computed automatically from the observed objective values
with a configurable safety margin, ensuring r is dominated by all
observed points along every objective.

\section{Recommendation Loop}

At every iteration k, the Bayesian optimization node executes the
following procedure inside BayBe's \emph{Campaign} API:

\begin{enumerate}
\def\labelenumi{\arabic{enumi}.}
\item
  The surrogate model is updated by refitting the Gaussian process from
  scratch on the full set of measurements \(\mathcal{D}_{k}\)
  accumulated so far (Section S.2).
\item
  The acquisition surface is constructed by instantiating qLogEI (single
  objective) or qLogNEHVI (multi-objective) with the updated posterior
  (Section S.3).
\item
  Optimization is done by maximizing the acquisition value over the
  unevaluated portion of the search space using a batch acquisition
  optimizer to obtain a candidate batch \(X_{k}\) of size q.
\item
  Each recommendation is cross-checked against the original search space
  (to enforce constraints and the substance label set) and against the
  historical ledger (to filter molecules already characterized in any
  prior iteration).
\item
  The validated \(X_{k}\ \)is forwarded to the appropriate
  characterization node, and the returned measurements are appended to
  \(\mathcal{D}_{k + 1}\), while the loop repeats.
\end{enumerate}

The de-duplication step in (4) is essential because the Gaussian process
posterior near previously observed points has very low variance, an
unconstrained acquisition optimizer would otherwise be free to recommend
arbitrarily small perturbations of already-tested compounds. Filtering
against the experiment ledger forces the recommended batch to occupy
genuinely new regions of chemical space, preserving the explore-exploit
trade-off that motivates the use of Bayesian optimization in the first
place.

\section{List of Properties Currently Supported by SABLE}

RDKIT\_PROPERTIES = \{

"molecular\_formula", "exact\_mass", "molecular\_weight",
"heavy\_atom\_count",

"ring\_count", "aromatic\_ring\_count", "h\_bond\_donors",
"h\_bond\_acceptors",

"rotatable\_bonds", "logp", "tpsa", "qed"

\}

STOPLIGHT\_PROPERTIES = \{

"alogp", "ampc\_aggregation", "cns\_activity", "cruzain\_aggregation",

"fsp3", "firefly\_luciferase", "hba", "hbd", "molecular\_weight",

"nano\_luciferase", "num\_heavy\_atoms", "num\_quaternary\_carbons",

"number\_of\_rings", "number\_of\_rotatable\_bonds",
"polar\_surface\_area",

"redox\_interference", "solubility\_water", "thiol\_interference"

\}

BOLTZ\_PROPERTIES = \{"binding\_affinity", "affinity"\}

\begin{table}
	\caption{Summary of different agentic systems chemistry and
their design approach}
	\centering
	\small
	\setlength{\tabcolsep}{3pt}
	\renewcommand{\arraystretch}{1.15}
	\begin{tabularx}{\textwidth}{@{}*{5}{>{\raggedright\arraybackslash}X}@{}}
		\toprule
System & Agent role & Candidate selection & Synthesizability constraints & Closed-loop with experiments \\
		\midrule
ChemCrow\cite{m.branAugmentingLargeLanguage2024a} & LLM as scientist &
LLM & Implicit & In theory \\
Coscientist\cite{boikoAutonomousChemicalResearch2023} & LLM as scientist
& LLM + heuristics & Implicit & Robotic synthesis \\
ChemReasoner\cite{sprueillChemReasonerHeuristicSearch2024} & LLM + MCTS
& MCTS-LLM & None & No \\
\textbf{SABLE (this work)} & LLM as orchestrator & Bayesian optimization
& Explicit (HEALER reaction templates + Enamine BBs) & Architected \\
		\bottomrule
	\end{tabularx}
	\label{tabs:tables1}
\end{table}

\begin{table}
	\caption{Example prompts and parsed outputs}
	\centering
	\small
	\setlength{\tabcolsep}{3pt}
	\renewcommand{\arraystretch}{1.15}
	\begin{tabularx}{\textwidth}{
  @{}
  >{\hsize=1.8\hsize\raggedright\arraybackslash}X
  >{\hsize=1.5\hsize\raggedright\arraybackslash}X
  >{\hsize=0.9\hsize\raggedright\arraybackslash}X
  >{\hsize=0.4\hsize\raggedright\arraybackslash}X
  >{\hsize=0.4\hsize\raggedright\arraybackslash}X
  @{}
}
		\toprule
Prompt & Seed Molecule & Targets & Iterations & Batch Size \\
		\midrule
"Optimize aspirin for better QED" & \seqsplit{CC(=O)Oc1ccccc1C(=O)O} & QED (max)
& 5 (default) & 10 (default) \\
"Find the analogs of ibuprofen with improved LogP and QED. Run for 10
iterations" & \seqsplit{CC(C)CC1=CC=C(C=C1)C(C)C(=O)O} & LogP (max), QED (max) & 10
& 10 (default) \\
"Starting from \seqsplit{O=C(NCc1ccccc1OC1CCC1)c1ccc2{[}nH{]}ncc2c1} Improve
binding affinity to Q92731" & \seqsplit{O=C(NCc1ccccc1OC1CCC1)c1ccc2{[}nH{]}ncc2c1}
& Binding affinity (min). Structure is fetched from UniProt & 5
(default) & 10 (default) \\
		\bottomrule
	\end{tabularx}
	\label{tabs:tables2}
\end{table}

\begin{figure}
	\centering
	\includegraphics[width=1\linewidth]{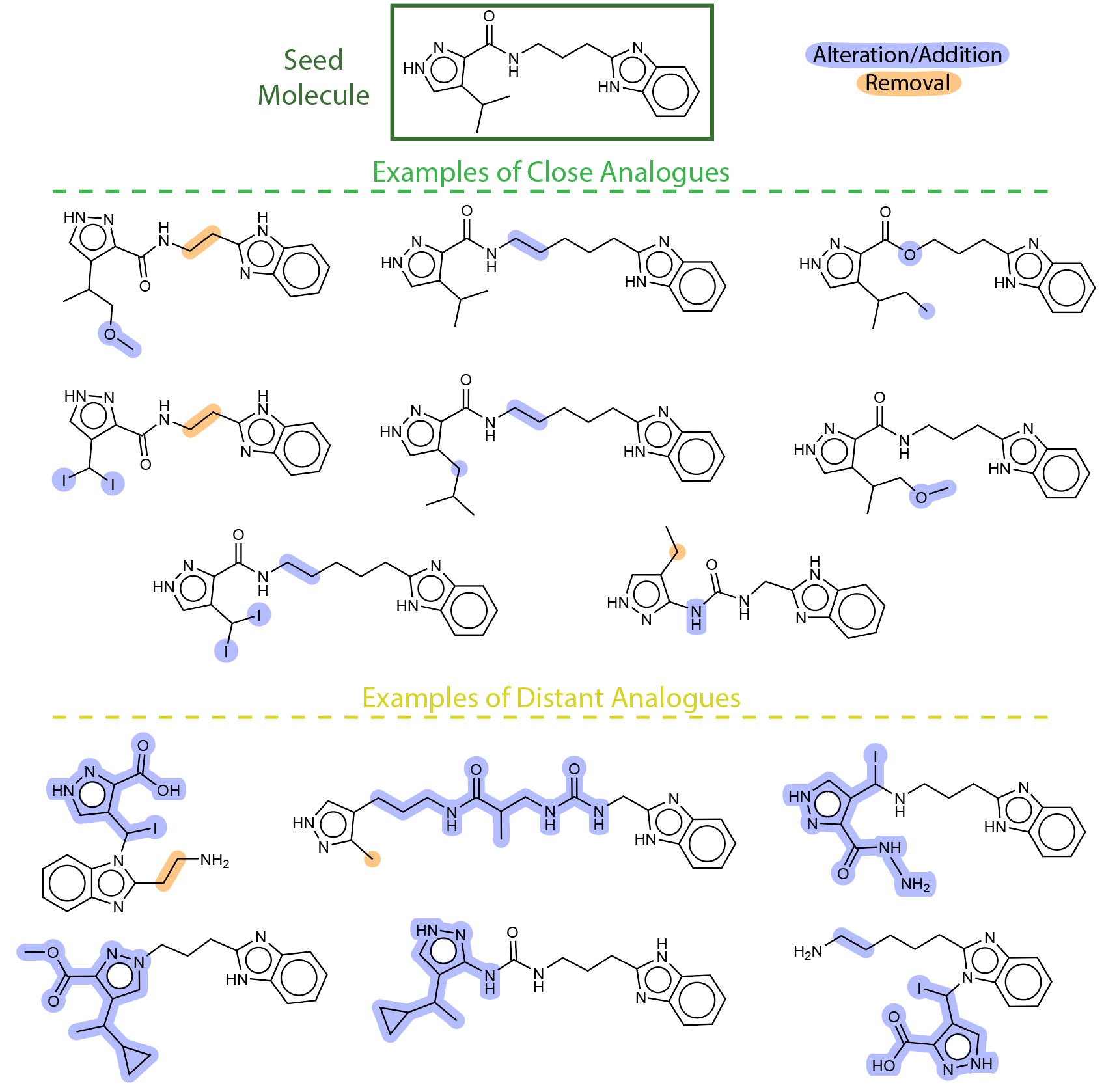}
	\caption{
 A sample starting compound and
enumerated analogs made by HEALER software. We show samples of close and
distant analogs based on the starting compound.}
    \label{figs:figs1}
\end{figure}

\begin{figure}
	\centering
	\includegraphics[width=1\linewidth]{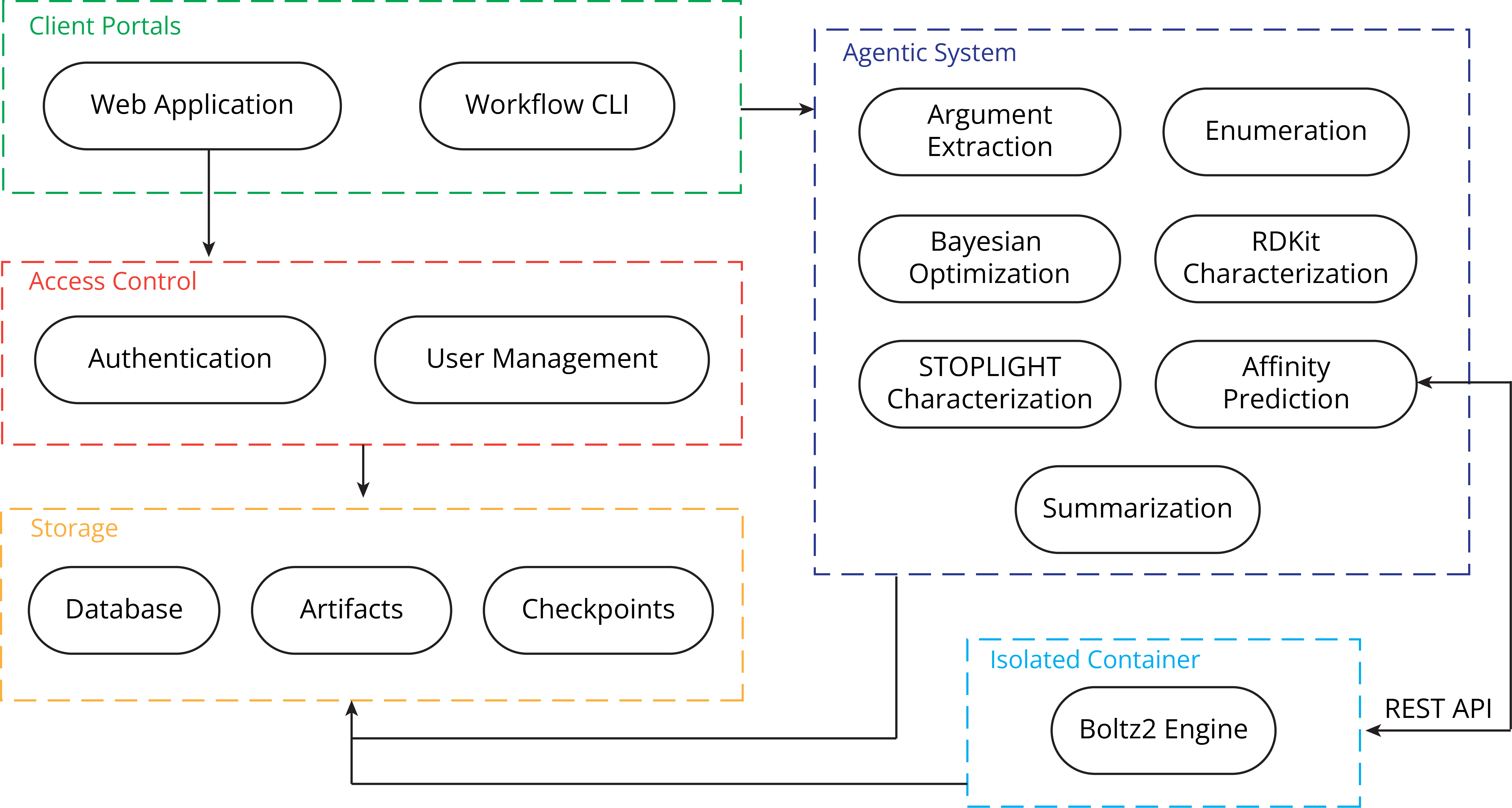}
	\caption{
Architecture of SABLE's agentic system. SABLE
features user-facing web and command-line interfaces through
authenticated client portals, with access control, user management,
persistent storage, artifacts, and checkpoints supporting reproducible
workflows. The agentic system coordinates task decomposition and
execution through argument extraction, molecular enumeration, Bayesian
optimization, QSAR/ADME/Tox characterization, affinity prediction, and
summarization nodes. Compute intensive calculations like affinity
predictions are containerized in a GPU-enabled server and communicate
with other nodes via a REST API.}
    \label{figs:figs2}
\end{figure}

\begin{figure}
	\centering
	\includegraphics[width=1\linewidth]{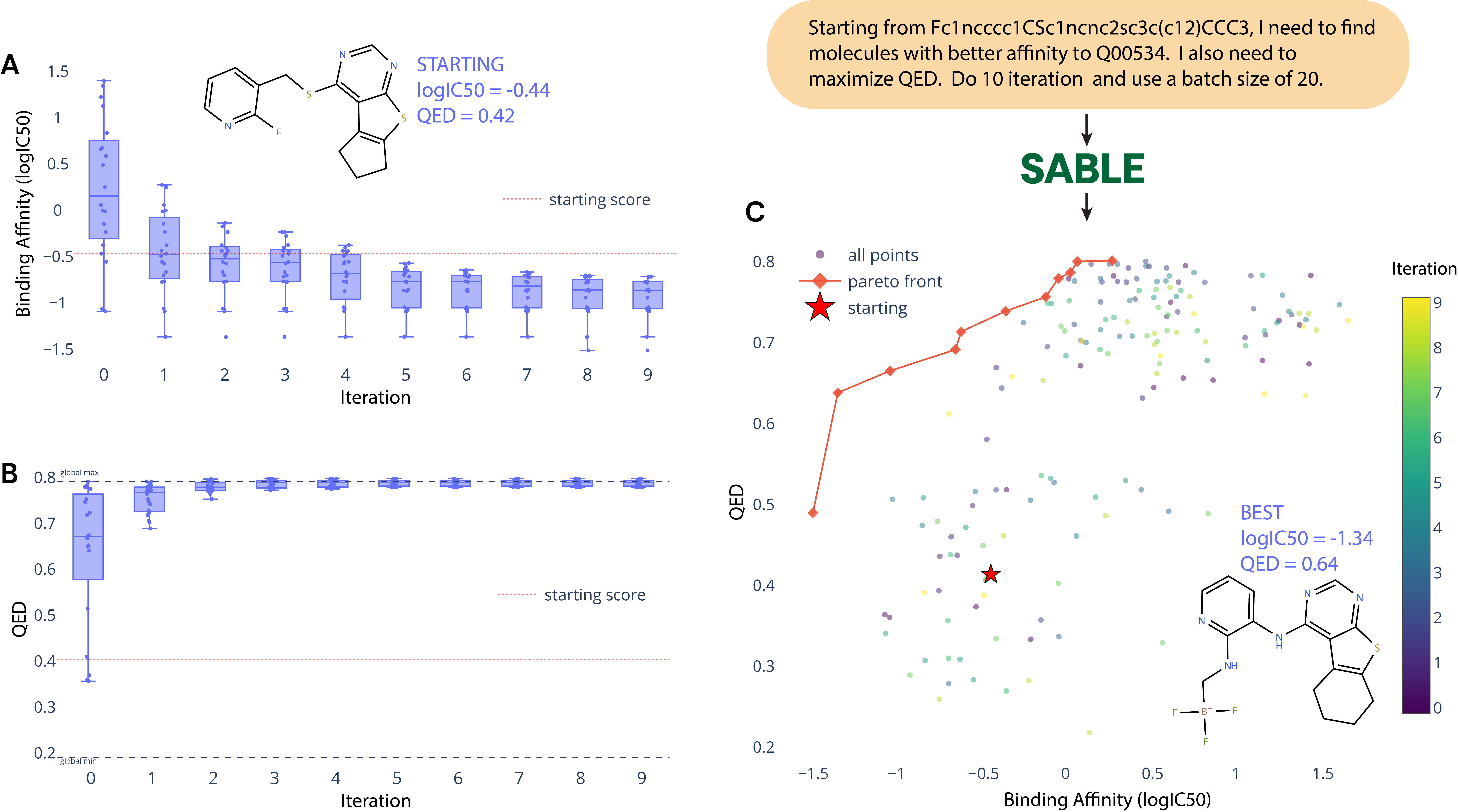}
	\caption{The dual-objective optimization of binding affinity
and QED. (A) Each boxplot shows the distribution of the
binding-affinity (logIC50) of the best 20 compounds selected and
characterized so far in the campaign. The structure of the starting
molecule with QED of 0.42 is shown. We find analogs that substantially
improve the binding affinity from -0.44 to -1.48 after eight (8)
iterations. (B) Iterative improvement of QED in the dual-objective
campaign with the starting compound shown in the red-dotted line. We
show that molecules with the best QED are found after the second
iteration of the campaign. (C) The Pareto frontier showing a trade-off
between binding affinity and drug-likeness. The best Pareto compound
that satisfied our trade-off with an affinity score of -1.34 and QED of
0.64 is shown.}
    \label{figs:figs3}
\end{figure}

\begin{figure}
	\centering
	\includegraphics[width=1\linewidth]{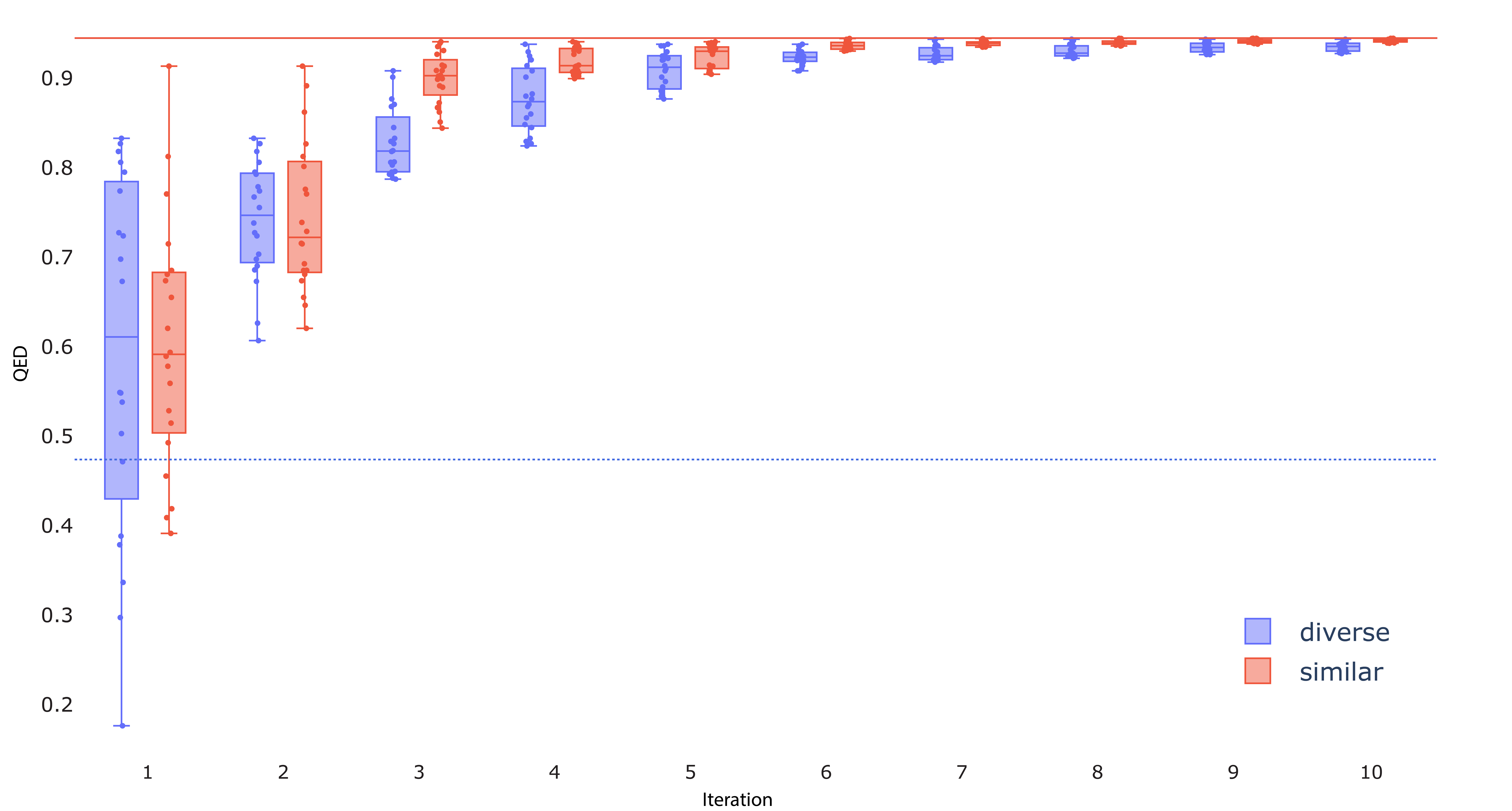}
	\caption{Optimization benchmark to determine the global
maximum in a 1 million compound size library. Boxplots show the
distribution of QED for the best 20 molecules obtained over ten
iterations. The search space generation used a mix of two strategies:
diverse and similar sampling of 1 million compounds. The blue dotted
line indicates the QED of the starting compound, while the red solid
line is the global maximum QED. The optimization routine finds the best
compound in three iterations for the similar search space and 7
iterations for the diverse search space.}
    \label{figs:figs4}
\end{figure}

\clearpage
\printbibliography[title={Supplementary References}]
\end{refcontext}
\end{refsection}

\end{document}